\documentclass[letterpaper, 10 pt, conference]{ieeeconf}% conference version

\IEEEoverridecommandlockouts% comment for final RAL version
\usepackage{amsmath}
\usepackage{amssymb}
\usepackage{array}
\usepackage{booktabs}
\usepackage{diagbox}
\usepackage{float}
\usepackage{graphicx}
\usepackage[hidelinks]{hyperref}
\usepackage{lipsum}
\usepackage{multirow}
\usepackage{siunitx}
\usepackage{times}
\usepackage{xcolor}

\renewcommand{\emph}[1]{\textit{#1}}
\newcommand{\revisedtext}[1]{#1}

\DeclareMathOperator*{\E}{\mathbb{E}}

\usepackage[acronym]{glossaries}
\newacronym{ae}{AE}{Autoencoder}
\newacronym{bc}{BC}{Behavioral Cloning}
\newacronym{cnn}{CNN}{Convolutional Neural Network}
\newacronym{ddpg}{DDPG}{Deep Deterministic Policy Gradients}
\newacronym{dof}{DoF}{Degrees of Freedom}
\newacronym{dqn}{DQN}{Deep Q-Network}
\newacronym{gan}{GAN}{Generative Adversial Network}
\newacronym{gps}{GPS}{Guided Policy Search}
\newacronym{mdp}{MDP}{Markov Decision Process}
\newacronym{ppo}{PPO}{Proximal Policy Optimization}
\newacronym{ransac}{RANSAC}{Random Sample Consensus}
\newacronym{relu}{ReLU}{Rectified Linear Unit}
\newacronym{rnn}{RNN}{Recurrent Neural Network}
\newacronym{rl}{RL}{Reinforcement Learning}
\newacronym{sgd}{SGD}{Stochastic Gradient Descent}
\newacronym{tanh}{tanh}{hyperbolic tangent}
\newacronym{tof}{TOF}{Time-of-Flight}
\newacronym{trpo}{TRPO}{Trust Region Policy Optimization}
\newacronym{vae}{VAE}{Variational Autoencoder}

\begin{document}

% Paper headers (only for final RAL version)
% \markboth{IEEE Robotics and Automation Letters. Preprint Version. Accepted January, 2019}
% {Breyer \MakeLowercase{\textit{et al.}}: Comparing task simplifications to learn closed-loop object picking}

% Paper title
\title{\LARGE \bf Comparing Task Simplifications to Learn Closed-Loop Object Picking Using Deep Reinforcement Learning}

% Paper authors
\author{Michel Breyer, Fadri Furrer, Tonci Novkovic, Roland Siegwart, and Juan Nieto%
%\thanks{Manuscript received: September, 10, 2018; Revised December, 13, 2018; Accepted January, 16, 2019.}%
%\thanks{This paper was recommended for publication by Editor T. Asfour upon evaluation of the Associate Editor and Reviewers' comments.}%
%\thanks{This work was supported in part by the Swiss National Science Foundation (SNF) through the National Centre of Competence in Research (NCCR) Digital Fabrication and the Luxembourg National Research Fund (FNR) 12571953. We would also like to thank Dario Mammolo  for his help with the robot experiments.}%
\thanks{M. Breyer, F. Furrer, T. Novkovic, R. Siegwart, and J. Nieto are with the Autonomous Systems Lab, ETH, 8092 Zurich, Switzerland, e-mail: \{michel.breyer, fadri.furrer, tonci.novkovic\}@mavt.ethz.ch, \{rsiegwart, nietoj\}@ethz.ch.}%
% \thanks{Digital Object Identifier (DOI): see top of this page.}%
\thanks{The accompanying video can be found at \url{https://youtu.be/ii16Zejmf-E}.}%
}%

% Make the title area
\maketitle

\pagestyle{empty}% comment for final RAL version
\thispagestyle{empty}% comment for final RAL version

\begin{abstract}

%% Motivation
Enabling autonomous robots to interact in unstructured environments with dynamic objects requires manipulation capabilities that can deal with clutter, changes, and objects' variability. 
%% Problem
This paper presents a comparison of different reinforcement learning-based approaches for object picking with a robotic manipulator. 
%% Approach
We learn closed-loop policies mapping depth camera inputs to motion commands and compare different approaches to keep the problem tractable, including reward shaping, curriculum learning and using a policy pre-trained on a task with a reduced action set to warm-start the full problem.
For efficient and more flexible data collection, we train in simulation and transfer the policies to a real robot.
%% Results
We show that using curriculum learning, policies learned with a sparse reward formulation can be trained at similar rates as with a shaped reward.
These policies result in success rates comparable to the policy initialized on the simplified task.
We could successfully transfer these policies to the real robot with only minor modifications of the depth image filtering.
%% Conclusions
\revisedtext{We found that using a heuristic to warm-start the training was useful to enforce desired behavior, while the policies trained from scratch using a curriculum learned better to cope with unseen scenarios where objects are removed.}

\end{abstract}

% Keywords
% \begin{IEEEkeywords}
% grasping, visual servoing, deep reinforcement learning, curriculum learning
% \end{IEEEkeywords}

%-------------------------------------------------------------------------------

\section{Introduction}

% \IEEEPARstart{I}{n} order for robots to assist us in everyday tasks, they need to be able to explore and interact with unstructured and dynamic environments found outside of traditional assembly lines and research labs.
In order for robots to assist us in everyday tasks, they need to be able to explore and interact with unstructured and dynamic environments found outside of traditional assembly lines and research labs.
Robust manipulation of objects is a key component in all robotic applications that require interaction with their surroundings. 
To perform a task, the robot needs to be able to perceive the environment through its sensors, plan and execute the next action, and at the same time handle noisy data, external disturbances and real-world uncertainties.
These challenges, together with the interdisciplinary nature of the problem, make this research area very complex.
Traditional approaches usually use methods from computer vision to interpret the sensor data and then use an analytic approach (policy) to plan the next action.
Manually designing policies that can cope with the complexity of the high-dimensional sensory input is difficult and often results in very tailored solutions to a given problem, which can be fragile to changes in the setup or task definition. 
Data-driven approaches, however, have shown to be powerful in such cases, given enough experience.
\gls{rl} is a general framework for training agents to acquire desired skills from trial-and-error by providing a reward for successful executions.
It is able to find a complex mapping from a high-dimensional input space to the desired actions, without the need to explicitly model this relationship.
However, depending on the complexity of the task, large amount of data might be required to learn the desired behaviour.
In particular, the exploration phase can take a long time in the presence of large and continuous state and action spaces.
To reduce the training time, the complexity of the manipulation task can be incrementally increased, thus allowing the learning algorithm to converge faster at each step.
Simulators can be used as a less expensive and faster alternative to real-world data collection.
However, transferring policies from simulation to the real world presents numerous difficulties, like sensor noise, contact physics approximations, simulated friction inaccuracies, etc., that can influence the final result on the real system.

\begin{figure}[t!]
  \centering
  \includegraphics[width=0.9\columnwidth]{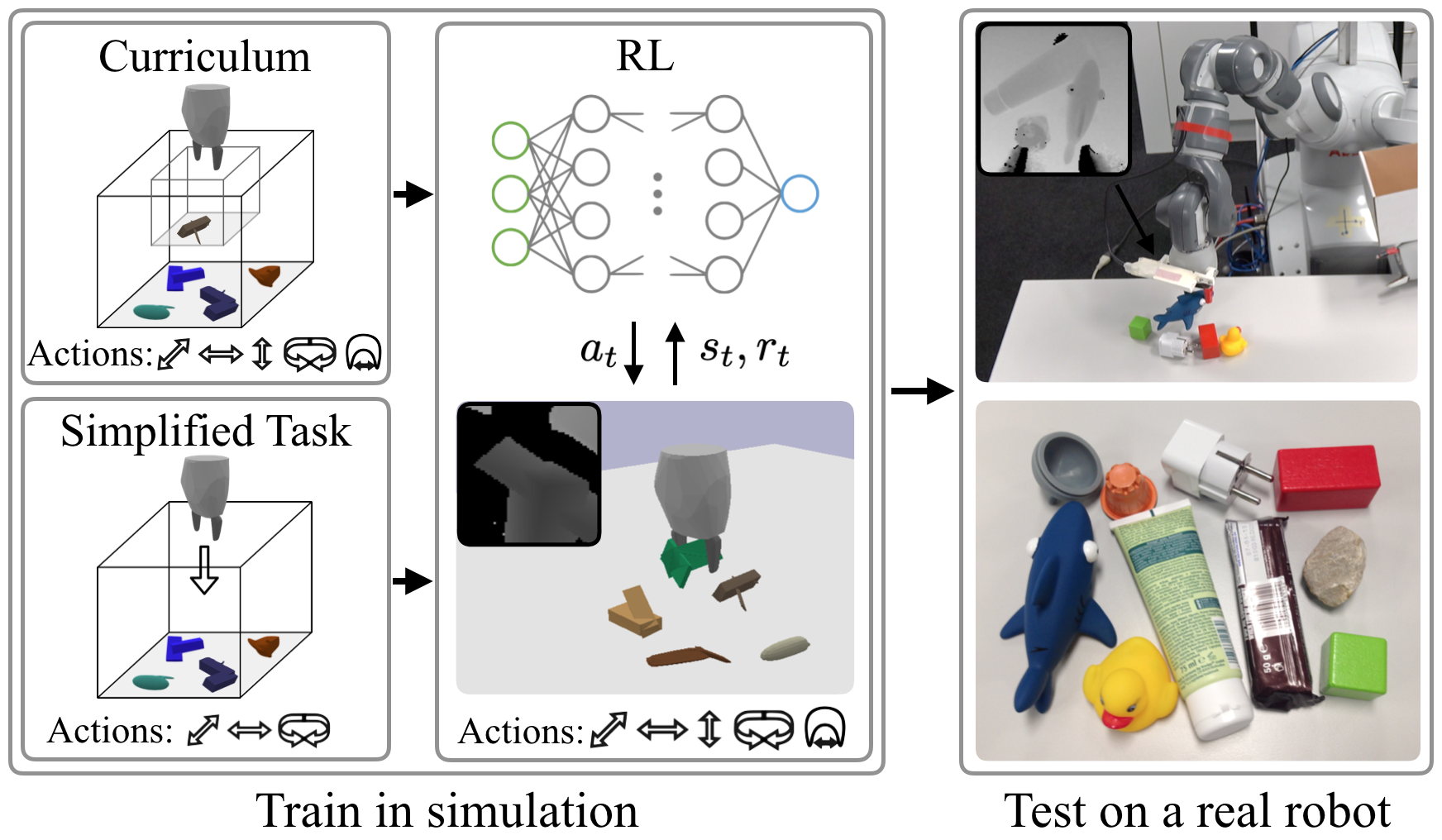}
  \caption{\revisedtext{We consider the problem of learning closed-loop policies for the combined task of reaching, grasping and lifting objects.
  The policies map images captured by a wrist-mounted depth camera to end effector motion and gripper opening and closing actions.
  We compare different approaches to improve the efficiency of training and performance of the final controllers, including reward function shaping, designing a curriculum of tasks with increasing difficulty, as well as using a partially scripted policy to provide a warm start for the full problem.
  Policies are trained in simulation and evaluated on a set of unseen objects, both in simulation and real-world experiments.}}
  \label{fig:teaser_image}
\end{figure}

In this work, we explore \gls{rl} approaches to train agents that interact with their environment in a fully closed-loop manner in order to maximize future reward.
We learn policies for the full task of reaching, grasping and lifting, which map depth images captured from a wrist-mounted camera to end effector displacements and gripper actions of a robotic arm, without relying on heuristics for the grasp decision.
We explore different mechanisms to reduce training costs.
First, we separate perception and control by learning a compressed image representation using the latent space of an autoencoder.
Second, following the methodology of curriculum learning~\cite{bengio2009curriculum}, we guide the training of our models by progressively increasing the workspace with the agent's performance and compare this method against reward function shaping and using a heuristic to bootstrap the full problem.
Finally, the entire training is performed in simulation and we report the required adjustments and findings from transferring policies to a real platform.
In summary, the contributions of this paper are:

\begin{itemize}
  \item A closed-loop end-to-end formulation for the combined task of reaching, grasping and lifting different objects.
  \item A case study of applying curriculum learning to guide training on this challenging task, including a comparison to alternative approaches.
  \item A presentation of findings from transferring policies learned exclusively in simulation to a real-world table clearing task.
\end{itemize}

%-------------------------------------------------------------------------------

\section{Related Work}\label{sec:related_work}
The task of reaching, grasping, and lifting can be solved using a large variety of approaches.
First, we present a set, which we think is representative of general methods to solve this problem.
In the second part, we highlight a selection of \gls{rl} formulations, their robotic applications, and how they are used to partially or completely solve this manipulation task.

\emph{Grasp planning} considers the problem of detecting grasp candidates that maximize the probability of success for a given environment and gripper configuration.
Early approaches relied on geometric reasoning, often assuming knowledge of the shape and physical properties of the involved objects~\cite{nguyen1988constructing, shimoga1996robot}.
Data-driven approaches on the other hand aim at learning models from labeled data that can exploit visual cues and generalize to unseen objects~\cite{bohg2014data}.
Lenz et al.~\cite{lenz2015deep} trained a deep neural network on a small set of human-labeled images to predict the success of grasps on novel images.
A different approach is to exploit analytic grasp theory to generate labeled data from synthetic point clouds~\cite{gualtieri2016high, mahler2017dex, viereck2017learning} while a third line of research learns their models in a self-supervised manner directly from physical trials~\cite{pinto2016supersizing, levine2016learning}.
Compared to the first two classes, the self-supervised approaches do not require any prior knowledge on grasp theory or human labeled samples.
While some of the mentioned works improve robustness by iteratively recomputing the best grasp configuration~\cite{levine2016learning, viereck2017learning}, they do not consider the long-term consequences of actions required to learn more complex behaviors.
To enable an agent to learn such action sequences, one can pose this task as a \gls{rl} problem.

\emph{Reinforcement learning}~\cite{kaelbling1996reinforcement, sutton1998introduction} is a general framework that considers autonomous agents who learn to choose sequences of control decisions that maximize some long-term measure of reward.
To tackle the high-dimensional and continuous problems typically found in robotic applications, early works relied on task-specific, hand-engineered policy representations~\cite{peters2008reinforcement, stulp2012reinforcement}.
Combining \gls{rl} with the expressive power of deep neural networks has lead to some impressive results in various complex decision making problems~\cite{mnih2015human,silver2016mastering}.
Due to the high data requirements, popular benchmarks often focus on video games~\cite{mnih2015human} and simulated control problems~\cite{lillicrap2015continuous, duan2016benchmarking}.
However, a number of works have applied \gls{rl} to real-world manipulation tasks.
One of the most notable ones is the Guided Policy Search~\cite{levine2016end}, which trains a large neural network policy in a supervised manner on samples collected with trajectory-based \gls{rl}.
Other works tackle individual skills, such as opening doors~\cite{gu2017deep} or lifting and stacking blocks~\cite{popov2017data}.
\revisedtext{Our problem formulation is closest to} Quillen et. al.~\cite{quillen2018deep}, who compared different off-policy \gls{rl} algorithms for bin-picking in clutter with a large set of training and unseen test objects.
This work got extended~\cite{kalashnikov2018qtopt} to include gripper actions and a decision variable on when to terminate an episode.
In the last two approaches the training data is generated from a scripted policy.
In contrast to their approaches, additionally to a heuristic policy initialization, we explore curriculum learning to make the problem tractable.

Sparse reward formulations are naturally suited for many goal-oriented manipulation tasks, but also create challenges leading to techniques such as augmenting reinforcement signals through reward shaping~\cite{gu2017deep, popov2017data}, learning from expert demonstrations~\cite{peters2008reinforcement, rajeswaran2017learning,pfeiffer2018reinforced} and \emph{curriculum learning}~\cite{bengio2009curriculum}.
The latter proposes to guide learning by presenting training samples in a meaningful order with increasing complexity and has been applied to supervised learning for sequence prediction~\cite{bengio2015scheduled} and \gls{rl} to acquire a curriculum of motor skills of an articulated figure~\cite{karpathy2012curriculum}.
\revisedtext{Akin to curriculum learning, Popov et al.~\cite{popov2017data} sample initial states along expert trajectories.}
Recent related work proposed to use a \gls{gan} to automatically generate goals of increasing difficulty~\cite{held2017automatic}, generate start state distributions that gradually expand from a given goal state~\cite{florensa2017reverse} \revisedtext{and training a teacher to automatically choose samples for the learner~\cite{matiisen2017teacher}}.
In our work, due to the large diversity of objects, goal states are not easily available. Therefore, our curriculum schedule increases both the space from which initial states are sampled as well as the final lifting height, where the target reward is awarded.

Opposed to using large-scale data collection on real robots~\cite{pinto2016supersizing, levine2016learning},~\revisedtext{\cite{kalashnikov2018qtopt}}, we perform training in simulation as a less expensive, faster and, safer alternative.
However, deploying policies learned in simulation on a real system requires to \emph{bridge the reality gap} induced by differences in sensing and dynamics, and is a very active field of research.
One approach is to close the gap by making the simulation match the real system as closely as possible through system identification~\cite{tan2018sim}.
Another approach is to expose the learning agent to a range of different environments through domain randomization,  forcing it to learn a robust representation that generalizes to the real-world~\cite{tobin2017domain, james2017transferring}.
Lastly, models can be adapted to new domains, e.g. by using progressive networks~\cite{rusu2016sim}, learning correspondences using a pairwise loss function~\cite{tzeng2015adapting} or using generative adversarial networks to map simulated images to realistic looking ones~\cite{bousmalis2017using}.
In this work, similarly to~\cite{johns2016deep} and~\cite{viereck2017learning}, we explore directly transferring trained models to the real world with only small modifications.
In contrast to their approaches, our policies also need to predict height displacements and grasp decisions which was found to be challenging.

%-------------------------------------------------------------------------------

\section{Approach}\label{sec:approach}

\subsection{Problem Formulation} \label{sec:problem_formulation}

We consider the combined task of reaching, grasping and lifting objects using a parallel-jaw gripper and a wrist-mounted depth camera.
Our goal is to find a closed-loop policy $\pi(a_t | s_t)$ through model-free \gls{rl} that maps sensor measurements to end effector displacement and gripper controls.
The input $s_t$ contains visual information captured from the depth camera and the current gripper opening width.
To keep model sizes of our policy small, we first learn a lower-dimensional encoding of the depth images, which is then concatenated with the gripper width.
This process is described in more details in Section~\ref{sec:agent_model}.
The 5-dimensional, continuous action vector 
\begin{equation}
  a_t =\begin{bmatrix} t_x & t_y & t_z & \psi & g \end{bmatrix}
\end{equation}
includes $xyz$-translation and yaw rotation $\psi$ of the robotic hand as well as the gripper opening width $g$.
The movement of the hand is performed relative to the gripper's frame, complementing our wrist-mounted camera setup.
The translation vector is clipped to a maximum length of \SI{1}{cm} per step, requiring many iterations to finish the task and allowing the agent to react to dynamic changes in the environment.
The gripper width command $g$ is interpreted as a binary decision, with negative/positive values being mapped to a closed/opened hand respectively.
In the reminder of this section, we are going to present the \gls{rl} training process, agent model and different approaches we explored for speeding up training, including reward shaping, curriculum learning and transfer learning.

\subsection{\revisedtext{Reinforcement Learning}}\label{sec:reinforcement_learning}

\subsubsection{\revisedtext{Background}}

Following~\cite{kaelbling1996reinforcement}, we model our \gls{rl} problem as a discrete time, finite horizon \gls{mdp} defined by the tuple $\left<S, A, r, \rho_0, p, T\right>$, where $S$ denotes the set of admissible states, $A$ the set of valid actions, \revisedtext{$r : S \times A \times S \to \mathbb{R}$} a real-valued reward function, $\rho_0$ the initial state distribution and \revisedtext{$p : S \times A \times S \to \mathbb{R}$} the (unknown) transition probability distribution.
At each time step $t$, an \gls{rl} agent observes the current state $s_t \in S$ of its environment and decides to take an action $a_t \in A$ according to a parameterized policy $\pi_\theta(a_t | s_t)$.
The execution of this action causes the system to transition to a new state $s_{t+1}$ according to the system dynamics $p$ and the agent receives a reward \revisedtext{$r(s_t, a_t, s_{t+1})$}.
\revisedtext{Episodes are terminated after a fixed number of steps $T$ or once a defined terminal state is reached.}
The goal of \gls{rl} is to \revisedtext{find parameters $\theta^*$ that maximize} the return $J(\theta){=}\E \left[\sum_{t=0}^{T-1} \gamma^t r(s_t, a_t, s_{t+1}) \right]$, where $\gamma \in [0,1]$ denotes a discount factor and the expectation is computed over the distribution of all possible trajectories \revisedtext{$\tau= \{s_0, a_0, \ldots, s_{T} \}$ with probabilities $p_\theta(\tau) = \rho(s_0) \prod_{t} p(s_{t+1} | s_t, a_t) \pi_\theta(a_t | s_t)$}.

\subsubsection{\revisedtext{Training Process}}\label{sec:training_process}

In our task of object picking, at the beginning of each episode, we sample from the initial state distribution by randomly selecting $n$ objects and placing them at a random pose within a workspace of size $l \times l$ on a flat surface, \revisedtext{where $l$ is the extent of the workspace}.
The number of objects $n$ is uniformly chosen between $1$ and $n_\text{max}$ for every new episode and the robot hand is placed pointing downwards at the center of the workspace with a distance $h_\text{robot}$ between its finger tips and the surface.

We consider the outcome of an episode as a success and terminate if, within the time horizon of $T{=}150$ control steps, \revisedtext{any} object was lifted for $h_\text{lift}$.
A natural reward function of this task would be a binary $r{=}r_T$ in case of success and $r{=}0$ otherwise. Such sparse rewards are difficult to learn from, requiring significant exploration.
For this reason, in order to guide training, we also consider an alternative shaped reward formulation, in which the agent additionally receives intermediate reward signals for lifting objects,
\begin{align} \label{eq:reward}
  r = (\text{grasp\_detected}) \cdot (r_g + c \cdot \Delta h),
\end{align}
with $r_T{=}10$, $r_g{=}1$, $c{=}1000$, and $\Delta h$ the difference in the robot's height since the last step. The first term in the equation is a binary function that returns $1$ if a grasp was detected and $0$ otherwise. Grasp detection is achieved by checking if the fingers stalled after a closing command was issued.
We also include a time penalty of $-0.1$ and $-(r_g + c \cdot \Delta h_\text{max})$ for the sparse and shaped reward functions respectively, where $\Delta h_\text{max}$ is the maximum allowed change in height per step. The latter is chosen such that rewards are shifted to negative values encouraging the agent to complete the task as quickly as possible.

\subsection{Workspace Curriculum}

Limited prior knowledge allows model-free \gls{rl} to be applied to a large set of tasks, but also renders exploration of interesting parts of the state space challenging. Particularly in manipulation tasks with large workspaces, the agent might waste significant training time exploring free space away from the objects. For this reason, following the formalism of Bengio et al.~\cite{bengio2009curriculum}, we propose a curriculum of workspaces with increasing sizes to guide training of our agents. 

Consider a sequence of training distributions, where the extent of the workspace $l$, initial robot height $h_\text{robot}$, target lift distance $h_\text{lift}$, and maximum number of objects $n_\text{max}$ \revisedtext{each} increase \revisedtext{linearly within a defined range} with a variable~$\lambda$, $0 \leq \lambda \leq 1$.
\revisedtext{A value of $\lambda = 0$ is mapped to the smallest possible value of each parameters and $\lambda = 1$ is mapped to their maximum value.
For $n_\text{max}$, we rounded to the nearest integer.}
The curriculum step $\lambda$ is increased step-wise each time the success rate averaged over a window of recent episodes reaches a certain threshold $\epsilon$.
This ensures that \revisedtext{the agent explores the state space close to the objects of interests in early stages of training} while allowing to scale to large workspaces in later stages.

\subsection{Transfer Learning}\label{sec:transfer_learning}

For comparison, we also consider agents pre-trained on a simplified task formulation, similar to~\cite{quillen2018deep}, that includes a heuristic to guide training.
Robot arm movements are restricted to $xy$-translation and yaw rotation with the $z$ component fixed to a constant downward movement.
Furthermore, the grasp decision is replaced by a heuristic that attempts a grasp once a given height threshold is reached.
The reward function for this task is binary and equals to $r_T$ if an object was successfully lifted for \SI{5}{cm}, $0$ otherwise.
Removing two \gls{dof} significantly decreases the complexity of the task, but also limits the behavior of the learned policies.
State-action pairs collected by \revisedtext{executing} an agent trained on this task can be augmented to be compatible with the original action description. \revisedtext{Given this data, we use \gls{bc} to train a policy predicting the full action space.
The weights of this policy provide a warm start for further fine-tuning through \gls{rl}.}

\subsection{Agent Model}\label{sec:agent_model}

\begin{figure}
  \centering
  \includegraphics[width=0.9\columnwidth]{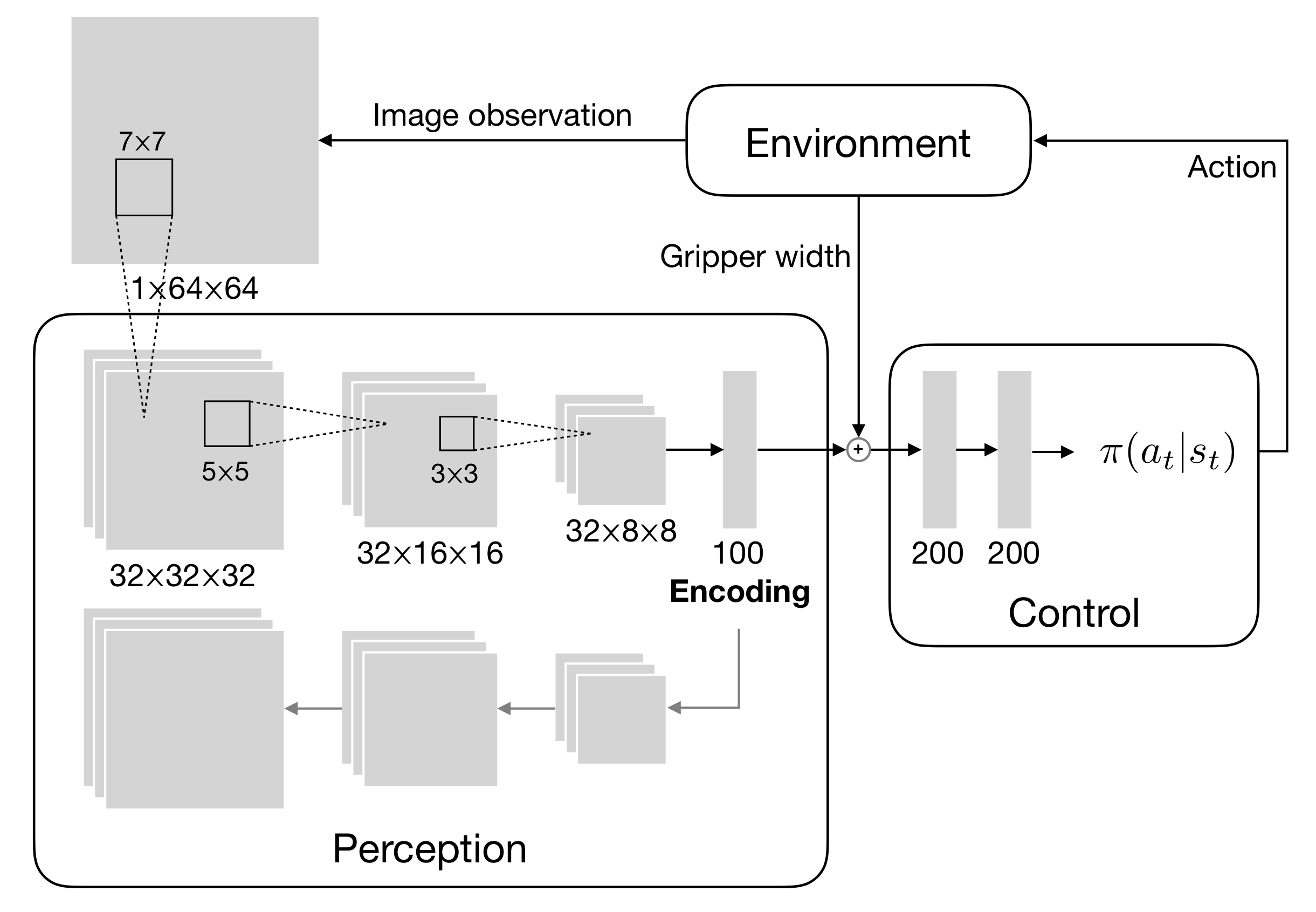}
  \caption{Overview of the two networks used in this work. At each time step, the agent receives an image of its environment. The \emph{perception network} generates a small-dimensional encoding, which is then concatenated with the current opening width of the gripper and passed to the \emph{control network} to determine the next action. The former is trained in unsupervised manner and kept fixed throughout training of the control policy.}
  \label{fig:agent_model}
\end{figure}

We separate the visual sensory and decision-making components of our agents. In a first stage, a perception network is trained to map image observations to a small-dimensional latent vector in an unsupervised manner. This network is then kept fixed and used to train a smaller policy to maximize the reward function described in~\ref{sec:reinforcement_learning}. Details of the different network architectures are depicted in Figure~\ref{fig:agent_model}.

\subsubsection*{Perception Module}

\begin{figure}
  \centering
  \includegraphics[width=0.7\columnwidth]{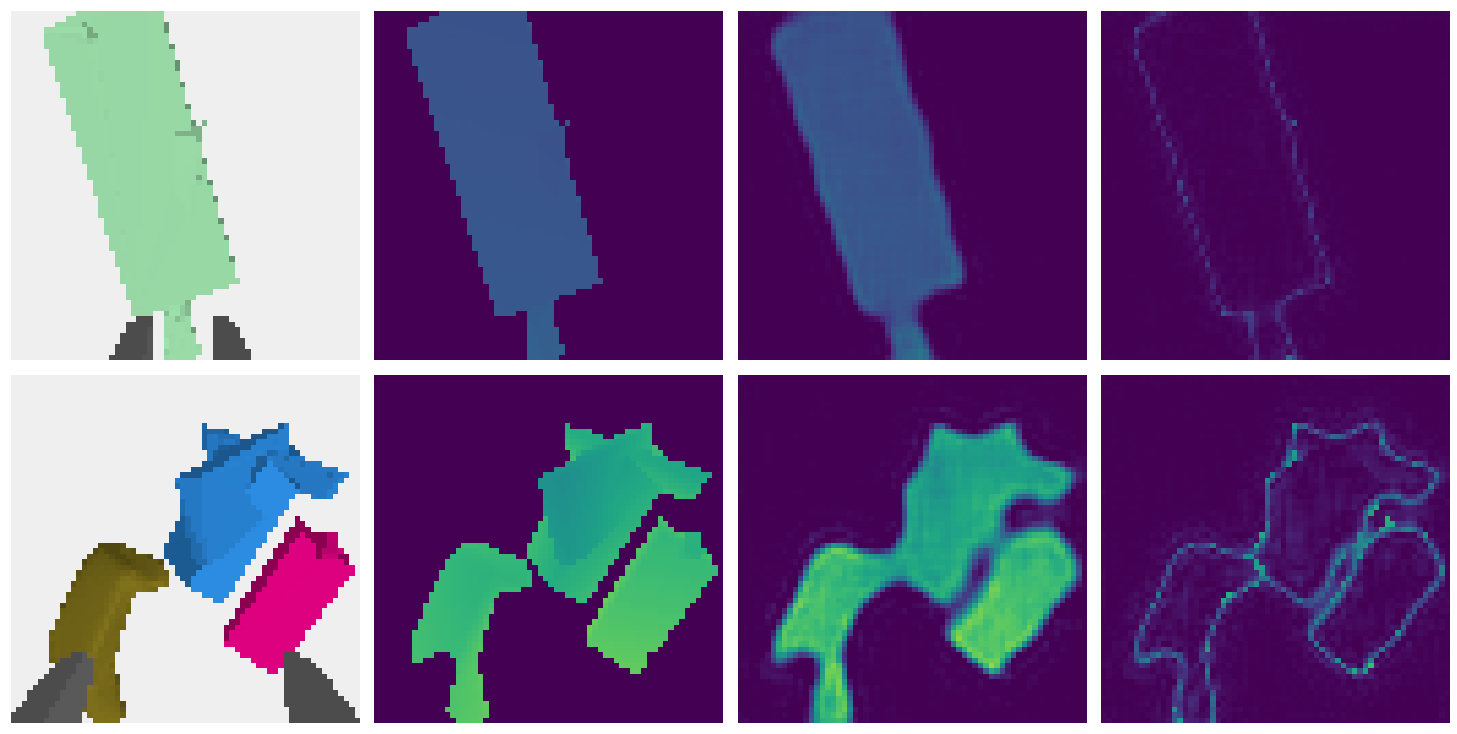}
  \caption{Two samples of images processed by our perception pipeline. From left to right, we show simulated RGB and filtered depth images, reconstructions produced by the autoencoder and the difference between originals and reconstructions.}
  \label{fig:encoder_results}
\end{figure}

The goal of the perception network is to encode information about the shape, scale and distance of the objects in the scene into a low-dimensional latent vector $z$. In this work, we use a simple autoencoder. The encoder consists of 3 convolutional layers followed by a fully-connected layer using leaky ReLU non-linearities. The decoder mirrors the architecture of the encoder to reproduce a full-sized image. Using a low-dimensional bottleneck and training the parameters to minimize the L2 distance between the original and reconstructed images forces the encoder to learn a compressed representation of the input. The training set was collected by running a random policy on the simplified task described in the previous section. Since they are not relevant for our task, we filtered out the plane and gripper fingers from the images. Figure~\ref{fig:encoder_results} shows two examples of original, filtered, reconstructed and error images using an encoder trained on a  dataset of 50000 images, using 120 epochs of the Adam optimizer~\cite{kingma2014adam}, with a learning rate of $2\cdot 10^{-4}$, and batch size of $128$. The same encoder weights were used throughout all experiments in this work.

\subsubsection*{Control Module} 

We use a small network that is trained separately from the perception network to map encoded observations to optimal actions.
Policies are modeled as multivariate Gaussian distributions.
A feed-forward neural network with two hidden layers and ReLU activations maps observations to the means of the distribution while the log-standard deviations are parameterized by a global, trainable vector.
Actions are normalized to the range of $[-1, 1]$ using a $\tanh$ output non-linearity.
Policy weights are optimized using \gls{trpo}~\cite{schulman2015trust}, a policy gradient method that performs stable updates by enforcing a constraint on the maximum change in policy distributions between two updates.

\subsection{Simulation}

Collecting data using a dynamic simulation and synthetic depth images instead of a real system has several advantages: it is faster, scales better, has lower cost, there is no need for supervision, automatic reset of experiments is easy to implement, and full state information is available.
For this reason, we focused on performing all training in simulation.
We constructed a virtual world using the Bullet physics engine~\cite{coumans2018} and added a disentangled robot hand whose position is controlled via a force constraint, avoiding the computation of inverse kinematics.
A virtual camera rendering $64 \times 64$ images was placed to match the viewpoint of the real setup. Depth images were generated using a software-renderer bundled with the physics engine and filtering was performed using masks provided by the engine.

\subsection{Transfer to the Real Platform}

We explore transferring policies trained in simulation to the real world without any fine-tuning of the network weights.
Ideally, images captured from the real camera would only need to be resized and cropped to match the dimensions of the simulated camera and then be passed into the encoder.
However, due to imperfect data and high noise levels, especially at the operating boundaries of the real sensor, some additional filtering was required.
In particular, we noticed increasing noise and some curvature towards image boundaries, as well as high noise around the gripper's fingers.
For this reason, we applied an additional elliptic mask to filter out the borders and dilated masks of the gripper's fingers.
The surface was detected and filtered using a \gls{ransac}~\cite{fischler1981random} based approach.

%-------------------------------------------------------------------------------

\section{Evaluation}\label{sec:evaluation}

The goal of our experiments is to evaluate and compare training times and final performance of the proposed models, as well as assess their capability to react to dynamic changes and transfer to the real world.

\subsection{Experimental Setup}

The platform used for evaluation consists of a position controlled 7-\gls{dof} arm of an ABB Yumi with a maximal payload of \SI{250}{g}.
The fingers of a stock gripper with opening width of \SI{5}{cm} were rubber-covered for better grip and reducing reflection.
A CamBoard pico flexx time of flight camera was attached to the wrist of the robot at a tilt angle of \SI{15}{\degree} as seen in the top right image of Figure~\ref{fig:teaser_image}.
In simulation, we used a model that matches the real robot and step the dynamics simulation with a size of \SI{5}{ms} which provided plausible physical behavior.
Training was performed on a set of procedurally generated random objects with diverse shapes\footnote{\url{https://sites.google.com/site/brainrobotdata/home/models}}.
Following \cite{quillen2018deep}, we split the dataset into 900 train and 100 test models and the objects were scaled to fit into the smaller gripper used in this work.
The grasping task was implemented on top of the OpenAI gym interface~\cite{brockman2016openai} and we based our implementation of \gls{trpo} on Rllab~\cite{duan2016benchmarking}.
Policy iterations were performed using a step size of $10^{-2}$ and a batch size of $10^4$ and $2 \cdot 10^4$ for the simplified/full task description respectively.
A curriculum of eight sets of workspace parameters was used with values linearly increasing in the ranges reported in Table~\ref{table:curriculum_parameters}.
\revisedtext{The curriculum step $\lambda$} is increased once a threshold success rate $\epsilon{=}0.7$ averaged over the last 1000 episodes was reached during training.

\begin{table}
  \caption{Parameter ranges for the curriculum sequences.}
  \centering
  \begin{tabular}{l|rr}
    \toprule
    {\bfseries Parameter} & {\bfseries Min. value} & {\bfseries Max. value} \\
    \hline
    $l$ & \SI{2}{cm} & \SI{20}{cm} \\
    $h_\text{robot}$ & \SI{4}{cm} & \SI{16}{cm} \\
    $h_\text{lift}$ & \SI{1}{cm} & \SI{10}{cm} \\
    $n_\text{max}$ & 3 & 5 \\
    \bottomrule
  \end{tabular}
  \label{table:curriculum_parameters}
\end{table}

\subsection{Simulated Experiments}

\subsubsection{Model comparison}

\begin{figure}
  \centering
  \includegraphics[width=\columnwidth]{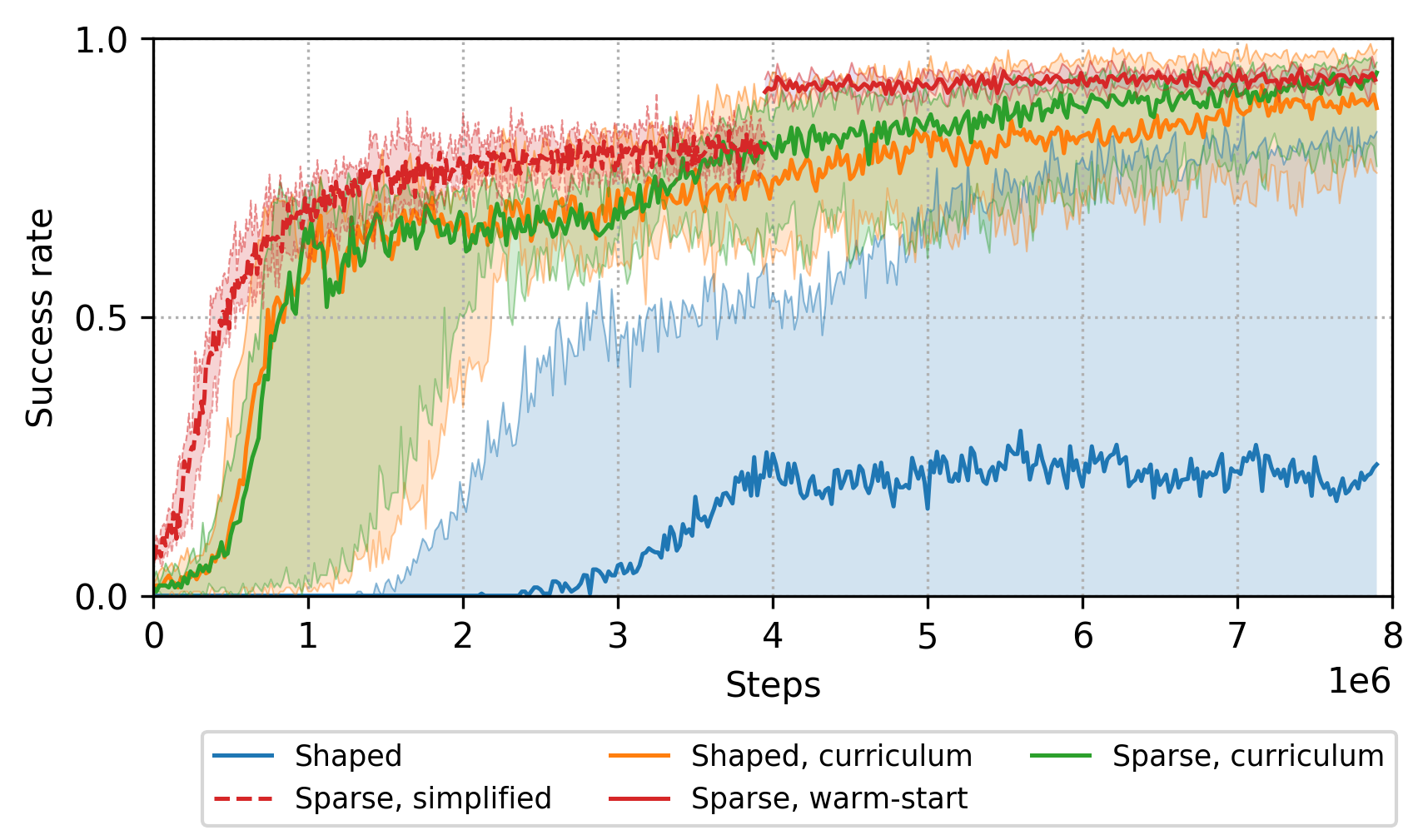}
  \caption{Learning curves for the different models analyzed in this work. Using a curriculum significantly speeds up training and leads to high final success rates that are comparable to the performance of an agent which was trained with a warm start provided by a grasping heuristic with fewer \gls{dof}.}
  \label{fig:model_comparison}
\end{figure}

\revisedtext{We analyze learning curves and the final performance of models trained on the full problem with only shaped rewards (\textit{shaped}), and using the proposed curriculum with both shaped and sparse reward formulations (\textit{shaped/sparse, curriculum}).
We also include agents trained on the simplified task (\textit{sparse, simplified}), the \gls{bc} (\textit{sparse, \gls{bc}}) and warm-started policies (\textit{sparse, warm-start}) described in Section~\ref{sec:transfer_learning}.}
Figure~\ref{fig:model_comparison} shows success rates of training iterations against \revisedtext{the number of environment interactions}.
For each model, we performed experiments with five different seeds and report the \revisedtext{median}, as well as the worst and best run, depicted as solid lines and shaded areas respectively.
Surprisingly, even when training on the full workspace from the start, the algorithm manages to reinforce the occasional intermediate reward provided when agents interact with objects.
However, results strongly varied over the seeds, with three out of the five runs failing altogether.
Using a curriculum significantly speeds up learning as well as the final performance of the agents.
We observe that the difference in the learning curves using the shaped and sparse rewards are quite small.
This confirms that providing easily reachable goal states at early stages of training acts as a mean of guiding the agent and speeding up the training process without artificially shaping the reward function.
Note that both of these models seem to stagnate temporally around \revisedtext{steps 1 to $3 \cdot 10^6$}.
This is due to the agents repeatedly reaching a success rate of $0.7$, triggering an increase in difficulty of the task until the curriculum parameters are set to their maximum values.
\revisedtext{This behavior is depicted in more details in Figure~\ref{fig:curriculum_study}(c), plotting the history of the curriculum step $\lambda$ along the history of success rates.}

\revisedtext{We can also observe that replacing two \gls{dof} with a heuristic (\textit{sparse, simplified}) results in a task that is considerably easier and faster to learn.
However, policies seem to converge to a lower success rate.
We found that a large fraction of these failure cases are due to a collision check, implemented to avoid the agent wasting time in case the robot stalled before reaching the low height threshold at which grasps are triggered.
This, combined with a larger time horizon allowing the agent to recover from failed grasp attempts, explains the jump in performance when continuing training using the full action set (\textit{sparse, warm-start}).}

\renewcommand{\arraystretch}{1.1}
\begin{table*}[t!]
  \centering
  \caption{Results of the simulated and real robot experiments.}
  \begin{tabular}{l | r | r | r r | r r | r r}
    \toprule
    \multirow{3}{*}{\bfseries Model} &
    \multicolumn{6}{c|}{\bfseries Simulation} &
    \multicolumn{2}{c}{\bfseries Real Robot} \\
    \cline{2-9}
    & 
    {\bfseries Single Object} &
    \multicolumn{1}{c|}{\bfseries Clutter} &
    \multicolumn{2}{c|}{\bfseries Table clearing (5)} &
    \multicolumn{2}{c|}{\bfseries Table clearing (10)} &
    \multicolumn{1}{c}{\bfseries Single Object} &
    \multicolumn{1}{c}{\bfseries Clutter}\\
     & success (\%) & success (\%) & success (\%) & \% cleared & success (\%) & \% cleared & success (\%) & success (\%) \\
    \hline
    Shaped & 74 & 80 & 54 & 62 & 64 & 59 & - & -\\
    Shaped, curriculum & 94 & 98 & 94 & 97 & 87 & 91 & 85 & 78\\
    Sparse, curriculum & 96 & 91 & 86 & 90 & 77 & 81 & 75 & 58\\
    Sparse, \gls{bc} & 54 & 56 & 38 & 35 & 35 & 22 & - & -\\
    Sparse, warm-start & 90 & 98 & 77 & 86 & 76 & 78 & 90 &70\\
    \bottomrule
  \end{tabular}
  \label{table:grasp_results}
\end{table*}

We evaluate the final performance of the agents over three different tasks: picking a singulated object, picking any object out of a pile of five objects and, similarly to the experimental setup in~\cite{mahler2017learning}, sequentially clearing objects from a flat surface until either all objects have been picked or the agent failed twice in a row.
Success rates are averaged over 200 episodes for the first two tasks or 40 sequences for the table clearing task using the \revisedtext{best performing} agent of each model.
For the latter, we additionally report the percentage of cleared objects.
Also, in order to investigate if our model generalizes to a larger number of objects than seen during training, we perform the table clearing task with an initial number of five and ten objects.
Comparisons are performed using the exact same sequence of object configurations and results are reported in Table~\ref{table:grasp_results}.
We can see that using our curriculum formulation reaches even slightly higher success rates than the warm-start model.
Generally, the latter performed well at properly aligning with objects, however the policies learned from scratch produced an interesting behavior, namely lifting the gripper after failed grasp attempts and in case that no object is within the current view, increasing chances of a successful grasp later in the episode.
Contrary, the warm-started policy presented a strong bias to move the gripper downwards, following the heuristic used to collect data for pre-training.
Pure \gls{bc} lead to poorer performance, mainly due to the agent failing to close the hand once it's aligned with the object.
Fine-tuning this policy with increased standard deviation for exploration quickly remedied this flaw.
\revisedtext{The combination of curriculum with the shaped reward function was found to be the most effective.}

Even though generally performance degrades, it is encouraging to see that the policies were able to cope with the larger number of objects present in the second table clearing experiment.
\revisedtext{The policies were also found to perform well over a range of different initial heights.}
All policies are closed-loop and react to changes in the object configuration and external perturbations. We refer to the accompanying video for an example of this behavior.

\subsubsection{\revisedtext{Ablation study of the curriculum parameters}}

\begin{figure}
  \includegraphics[width=\columnwidth]{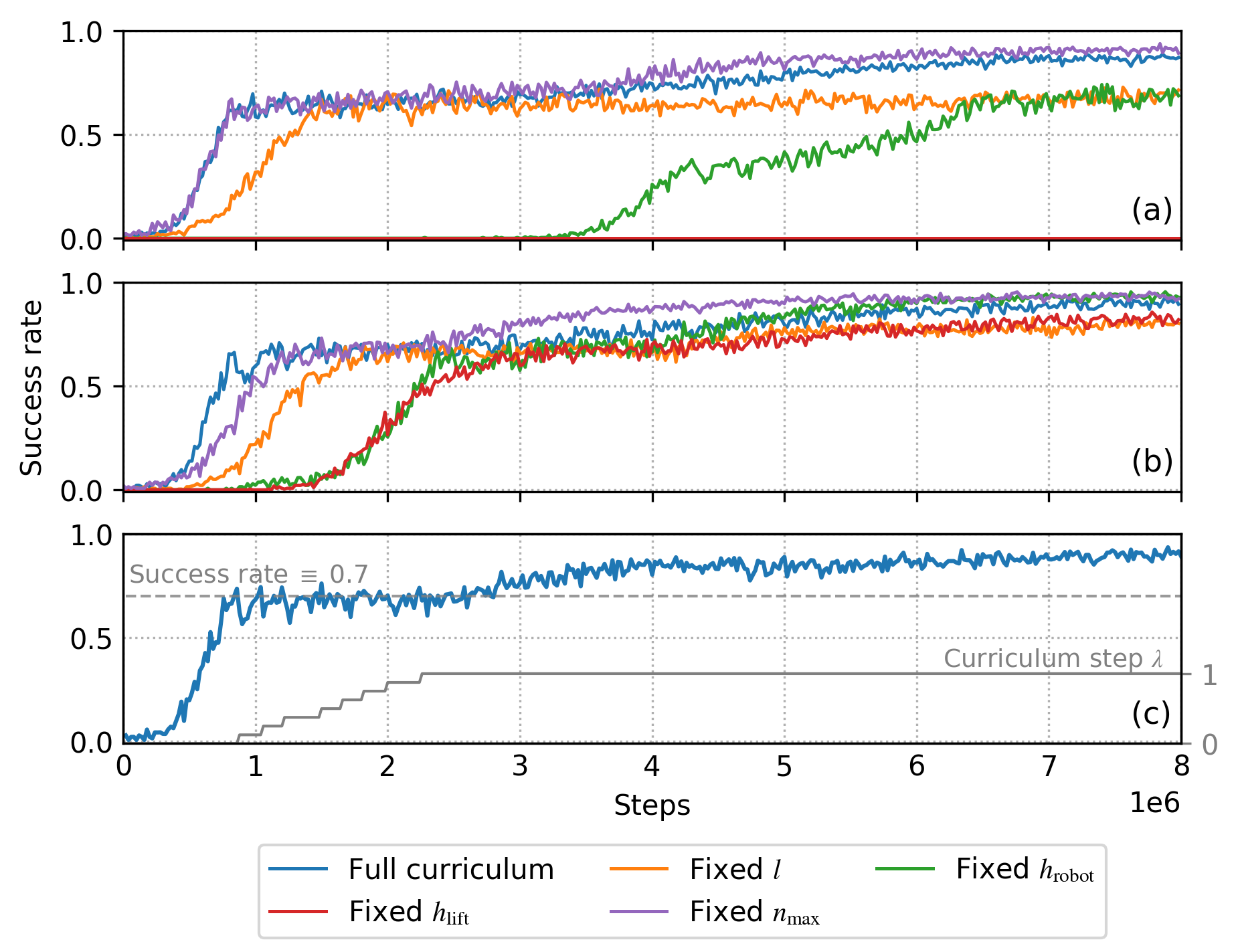}
  \caption{\revisedtext{Ablation study of the workspace curriculum parameters. For each experiment, one parameter of the curriculum was kept fixed at its maximum value. Subfigures (a) and (b) show the learning curves using the sparse and shaped reward formulations respectively. Subfigure (c) shows the learning curve and the history of the curriculum step $\lambda$ vs training steps for one run.}}
  \label{fig:curriculum_study}
\end{figure}

\revisedtext{In order to analyze the importance of the individual parameters in the curriculum, we performed multiple experiments using both reward formulations, each time keeping one of the parameters fixed at its maximum value reported in Table~\ref{table:curriculum_parameters}.
Similarly to the previous model comparison, we performed 5 runs with different seeds for each setting and report the median learning curves in Figure~\ref{fig:curriculum_study}. 
Fixing $n_\text{max}$ led to very similar, if not slightly improved, results compared to the full setting.
This is not surprising, as more objects in the workspace increase chances of meaningful interaction.
Fixing $h_\text{lift}$ makes all runs fail in the binary reward case, as the probability of sequences that lead to final states becomes very small.
In the presence of intermediate rewards for lifting the object, we observe that training still converges, but to lower success rates.
Performing the entire training with a large initial robot height $h_\text{robot}$ results in slower convergence, but still results in similar success rates in the case of the shaped reward function.
Finally, having a large workspace extent $l$ has a surprisingly small effect on exploration, but leads to lower success rates in the long run.}

\subsection{Real-world Experiments}

\begin{figure}[t]
  \centering
  \includegraphics[width=\columnwidth]{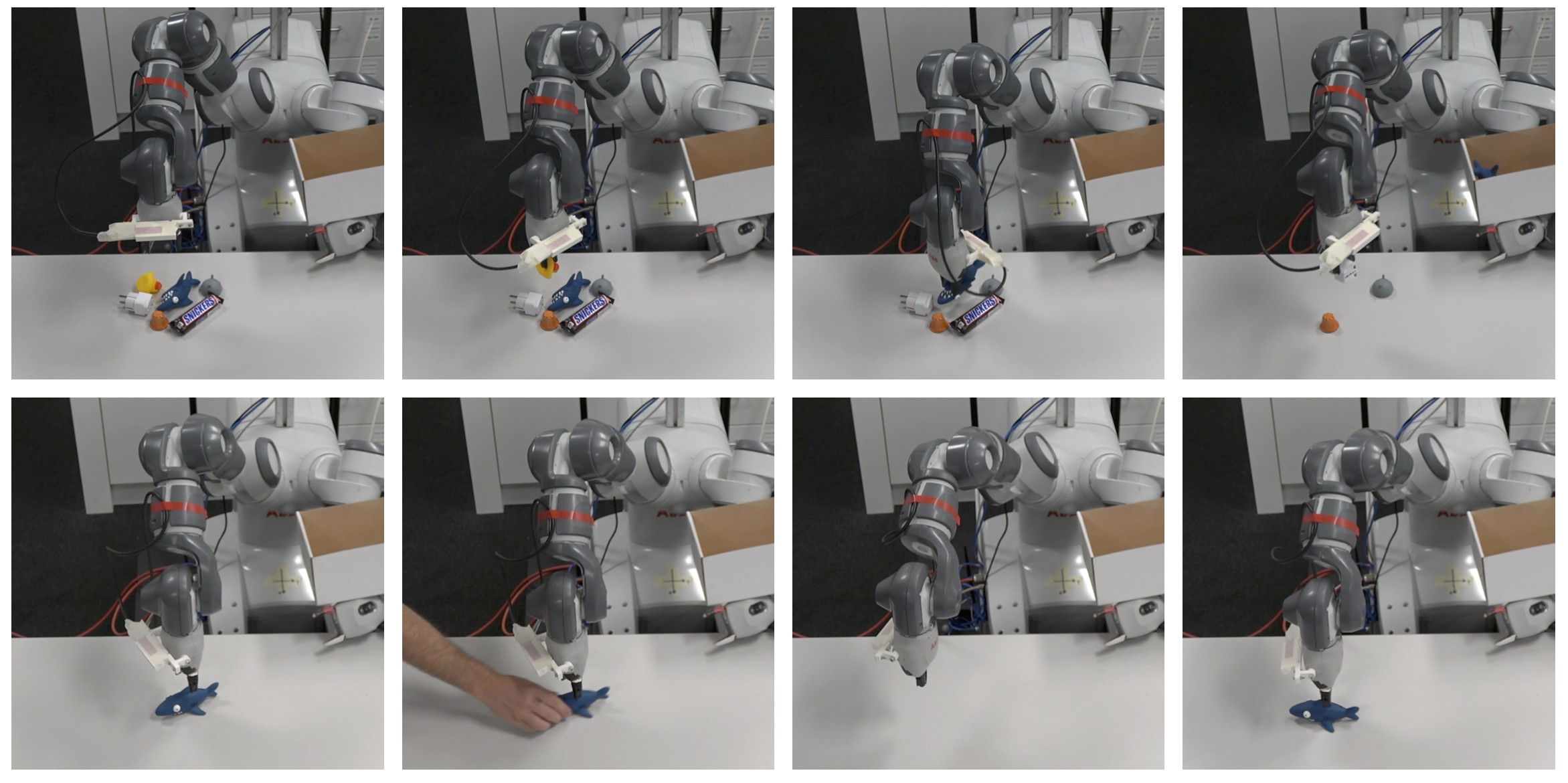}
  \caption{Two examples of our policy running on the real platform. The first sequence of images shows how the robot sequentially clears objects from the table. In the second example, shortly before the first grasp attempt, the object is removed from the scene. As a response, the agent lifts the robot arm searching for an object and realigns with the object as soon as it is placed back on the table before successfully terminating the episode.}
  \label{fig:grasp_sequence}
\end{figure}

To evaluate the transfer from simulation to the real system, we perform real robot experiments on a set of 10 unseen objects, \revisedtext{shown in Figure~\ref{fig:teaser_image}}. Experiments were conducted using the best run of the \textit{shaped} and \textit{sparse curriculum}, and \textit{sparse, warm-start} models, as they showed good performance in simulation.
Figure~\ref{fig:grasp_sequence} shows two sequences of the policy executed on a real robot.
We use the same singulated object and clutter picking experiments described in the previous section, with the addition of considering any action that leads the robot to halt, e.g. due to too high joint torques, as failures.
Objects are randomly placed in front of the robot by shuffling and placing the content of a box on a table and a total of 40 trials is performed for both tasks. \revisedtext{The results are shown in Table~\ref{table:grasp_results}}.
We observe a notable drop in performance compared to the simulated experiments, which is due to a couple of reasons.
First, high friction and approximate collision models used in simulation allowed some weak grasps, especially on the edges of objects, which fail in the real world.
Second, some collisions that occurred while the gripper was interacting with the objects, especially in the cluttered scenes, lead to the activation of safety mechanisms.
\revisedtext{In this regard, the \textit{sparse, warm-start} policy performed better than the \textit{sparse, curriculum} model, which we believe} is due to the collision check of the heuristic guiding the agent downwards, leading to zero reward.
Lastly, some runs failed \revisedtext{because of} the agent prematurely closing the fingers when approaching objects, \revisedtext{which can be explained by} the still existing differences between real and simulated images, \revisedtext{especially the high noise around the fingers}.
\revisedtext{Generally, the \textit{shaped, curriculum} model performed best, showing less collisions and more robust closing gripper decisions.}

\subsection{Discussion and Limitations}

Even though we observed worse performance on the real platform compared to simulated experiments, it is still encouraging that our policies achieved up to \revisedtext{$78 \%$} success rates in challenging picking tasks without any real robot data.
We are also convinced that these numbers can be improved by learning more robust policies in simulation, as explored in other works, either through randomizing various parameters of the dynamics~\cite{peng2017sim} and perception~\cite{tobin2017domain}, including some adversary applying disturbances to the system~\cite{pinto2017supervision, Pinto2017RobustAR} or by fine-tuning on the real platform.
Our translation actions result in jittery motions.
We expect policies trained to predict velocity or force actions to result in smoother trajectories.
The perception pipeline used in this work relied on the assumption that objects are placed on flat surfaces to perform the described filtering steps on the camera images.
However, this is not \revisedtext{always} given and could lead to a failure of our system.
\revisedtext{Considering the wrist-mounted camera placement, following~\cite{viereck2017learning}, we belief that this setup helps generalizing policies to different scenes}, since mostly the relative pose between objects and the gripper is of interest for choosing the next action.
However, sometimes it might be beneficial to rely on a top or over-the-shoulder view providing a better overview of the objects and scene around them.

%-------------------------------------------------------------------------------

\section{Conclusion}\label{sec:conclusion}

In this work, we presented a curriculum based approach to learn reactive policies for the task of object picking and compared this method against a formulation with shaped reward and cloning a heuristic with fewer \gls{dof}.
Curriculum learning allowed us to efficiently train policies using a natural sparse reward formulation and resulted in interesting behavior.
However, we also found that including prior knowledge in the form of heuristics can help enforcing desired behavior in a more direct way. The learned policies achieved high success rates in simulated picking tasks, both for single objects and in clutter. We also deployed agents learned in simulation to a real robot and reported our findings.

In future work, we would like to initialize agents with policies gathered through human generated actions in an augmented reality setting and imitation learning. Additionally, it would be interesting to investigate the benefits and/or drawbacks of our separated network approach compared to a single convolutional neural network policy in more details.
Finally, learning a hierarchy of policies for the different sub-tasks, e.g. reaching, grasping and lifting, might result in improved and more robust behavior.

%-------------------------------------------------------------------------------

\addtolength{\textheight}{-4.2cm}% balance the column lengths on the last page

%-------------------------------------------------------------------------------

% \section*{APPENDIX}

%-------------------------------------------------------------------------------

\section*{Acknowledgements}

We would like to thank Dario Mammolo  for his help with the robot experiments.
This work was supported in part by the Swiss National Science Foundation (SNF) through the National Centre of Competence in Research (NCCR) Digital Fabrication and the Luxembourg National Research Fund (FNR) 12571953.

%-------------------------------------------------------------------------------

\bibliographystyle{IEEEtran}
\bibliography{IEEEabrv,references}

\end{document}